%% file: aaai24.tex
\documentclass[letterpaper]{article} 
\usepackage{aaai24}  
\usepackage{times}  
\usepackage{helvet}  
\usepackage{courier}  
\usepackage[hyphens]{url}  
\usepackage{graphicx} 
\usepackage{natbib}  
\usepackage{caption} 
\usepackage{algorithm}
\usepackage{algorithmic}

\usepackage{newfloat}
\usepackage{listings}
\DeclareCaptionStyle{ruled}{labelfont=normalfont,labelsep=colon,strut=off} 
\floatstyle{ruled}
\newfloat{listing}{tb}{lst}{}
\floatname{listing}{Listing}
\usepackage{algorithm}
\usepackage{algorithmic}
\input{math_commands.tex}

\usepackage{booktabs}

\title{ANEDL: Adaptive Negative Evidential Deep Learning for Open-Set \\Semi-supervised Learning}
\author{
    Yang Yu\textsuperscript{\rm 1,*},
    Danruo Deng\textsuperscript{\rm 1,*},
    Furui Liu\textsuperscript{\rm 2},
    Qi Dou\textsuperscript{\rm 1},\\
    Yueming Jin\textsuperscript{\rm 3,\dag},
    Guangyong Chen\textsuperscript{\rm 2,\dag},
    Pheng Ann Heng\textsuperscript{\rm 1}
}
\affiliations{
    \textsuperscript{\rm 1}The Chinese University of Hong Kong, Hong Kong, China\\
    \textsuperscript{\rm 2}Zhejiang Lab, Hangzhou, China\\
    \textsuperscript{\rm 3}National University of Singapore, Singapore\\

}

\usepackage{bibentry}

\begin{document}

\maketitle

\begin{abstract}
Semi-supervised learning (SSL) methods assume that labeled data, unlabeled data and test data are from the same distribution. Open-set semi-supervised learning (Open-set SSL) considers a more practical scenario, where unlabeled data and test data contain new categories (outliers) not observed in labeled data (inliers). Most previous works focused on outlier detection via binary classifiers, which suffer from insufficient scalability and inability to distinguish different types of uncertainty. In this paper, we propose a novel framework, Adaptive Negative Evidential Deep Learning (ANEDL) to tackle these limitations. Concretely, we first introduce evidential deep learning (EDL) as an outlier detector to quantify different types of uncertainty, and design different uncertainty metrics for self-training and inference. Furthermore, we propose a novel adaptive negative optimization strategy, making EDL more tailored to the unlabeled dataset containing both inliers and outliers. As demonstrated empirically, our proposed method outperforms existing state-of-the-art methods across four datasets. Our code is avaiable: https://github.com/yuyang16101066/anedl. 

\end{abstract}

\section{Introduction}

\begin{figure*}[t]
	\centering
 \includegraphics[width=0.98\textwidth]{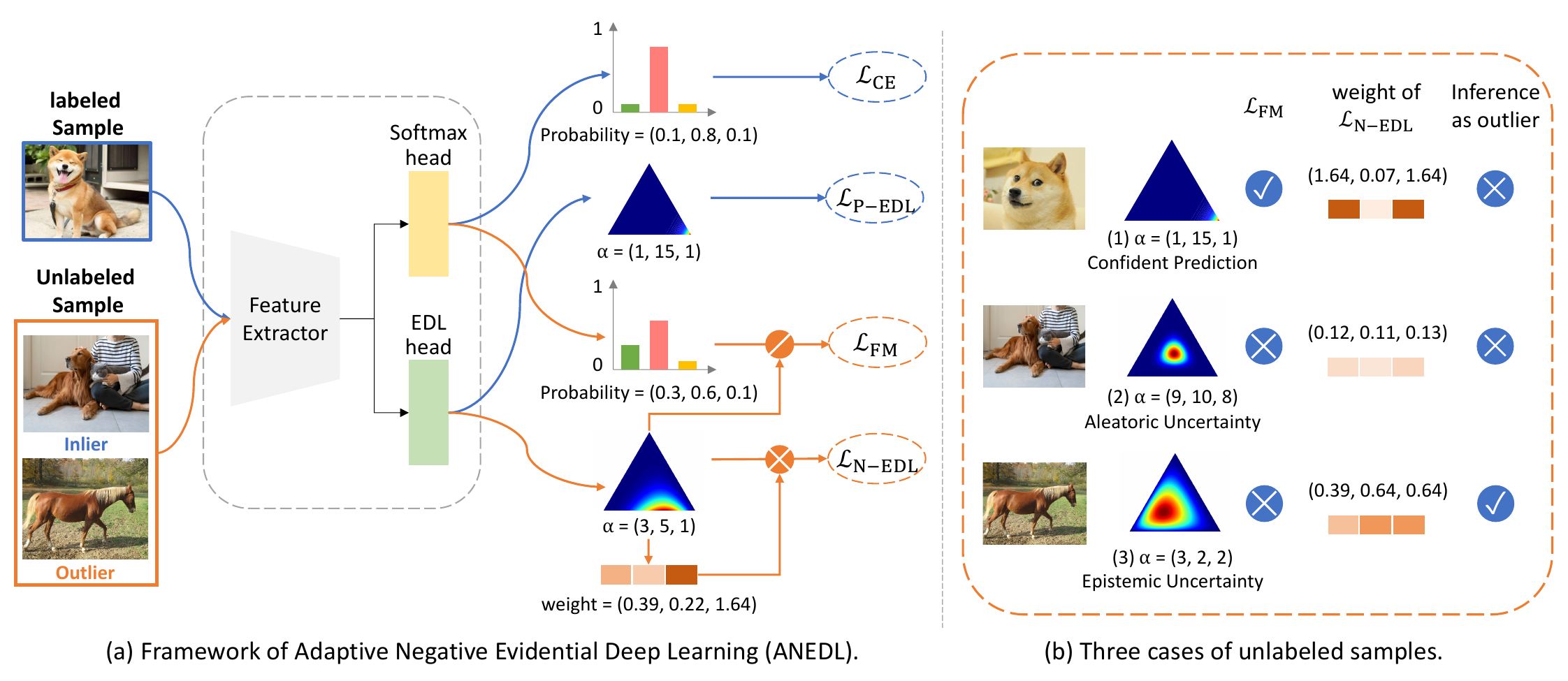}
	\caption{Overview of our proposed Adaptive Negative Evidential Deep Learning (ANEDL). (a) The framework of ANEDL consists of a shared feature extractor, a Softmax head and an EDL head. EDL head is used to detect outliers, while Softmax head is used to classify. To effectively leverage the information of inlier and outlier from unlabeled samples, we introduce Negative Optimization to explicitly regularize our EDL detector to output low evidence values for uncertain classes, while proposing adaptive loss weights to encourage the detector to pay more attention to these uncertain classes. (b) Three cases of unlabeled samples. Our model can quantify different types of uncertainty, including epistemic uncertainty due to lack of knowledge, and aleatoric uncertainty due to the complexity of samples from the distribution. Among them, only samples with confident predictions are used jointly with labeled samples to train the classification head.}
	\label{fig:anedl}
\end{figure*}

Semi-supervised learning (SSL) has recently witnessed significant progress by propagating the label information from labeled data to unlabeled data \cite{berthelot2019mixmatch, xu2021dash, wang2022freematch, zheng2022simmatch}. Despite the success, SSL methods are deeply rooted in the closed-set assumption that labeled data, unlabeled data and test data share the same predefined label set. In reality \cite{yu2020multi}, such an assumption may not always hold as we can only accurately control the label set of labeled data, while unlabeled and test data may include outliers that belong to the novel classes that are not seen in labeled data. During self-training, imprecisely propagating label information to these outliers shall interfere with model learning and lead to performance degradation. To this end, open-set semi-supervised learning (Open-set SSL) \cite{yu2020multi} has been emerging to tackle the problem. It first detects out-of-distribution (OOD) samples as outliers and then recognizes in-distribution (ID) samples as belonging to the classes from the predefined label set.

Existing Open-set SSL methods \cite{guo2020safe, chen2020semi} generally consist of an outlier detector and a classifier. To avoid propagating wrong label information to outliers during SSL training of the classifier, they first detect outliers and then optimize SSL loss only for unlabeled instances taken as inliers. To detect outliers, they aim to estimate confidence score of a sample being an inlier (reversely, uncertainty score for outliers). For example, T2T \cite{huang2021trash} proposes to detect outliers by cross-modal matching over binary detector.
OpenMatch \cite{saito2021openmatch} deploys a one-vs-all (OVA) classifier \cite{saito2021ovanet} as outlier detector to output the confidence score of considering a sample as an inlier and reject outliers with confidence score below a threshold for self-training in SSL. With outliers rejected, they can improve recognition accuracy of SSL methods.

We identify two main limitations in current state-of-the-art literature to tackle this task.
First, most previous works estimate confidence scores based on binary likelihoods generated from binary detectors activated by Softmax function. Such softmax-based networks only provide a point estimation of predicted class probabilities and 
cannot quantify whether the uncertainty stems from the lack of knowledge of outliers (epistemic uncertainty) or from the complexity of the samples within the distribution (aleatoric uncertainty) \cite{malinin2018predictive}. 
Moreover, when we tackle a K-way classification problem with a large K, the binary detectors are less robust to identify outliers from such a complex dataset that contains multi-class information \cite{carbonneau2018multiple}.
One advanced method, evidential deep learning (EDL) \cite{sensoy2018evidential} can explicitly quantify the classification uncertainty corresponding to the unknown class, by treating the network's output as evidence for parameterizing the Dirichlet distribution according to subjective logic \cite{josang2016subjective}. 
Compared with Softmax which only provides probabilistic comparison values between classes, EDL models second-order probability and uncertainty for each class.
In this work, we take the \emph{first} step to introduce EDL into the outlier detector. As shown in Figure \ref{fig:anedl}, EDL enables a multinomial detector to directly output 
evidence for each class, which can alleviate the insufficient scalability of existing binary detectors. Moreover, different types of uncertainties can be distinguished by the output vector,
so we propose to design different uncertainty metrics for self-training and inference.
Note that though EDL shows an impressive performance in uncertainty quantification, we empirically find its deficiency in learning representation preferred by classification tasks and consequent low performance in classification accuracy. Therefore, we propose to take advantage of both Softmax and EDL, by keeping Softmax for representation learning. 

The other limitation of prior methods is that they do not effectively leverage the information contained in unlabeled data. The detector should output low confidence scores for outliers, but this information is underutilized for model regularization, which may result in suboptimal performance \cite{malinin2018predictive}.
In this regard, we propose \emph{adaptive negative optimization} tailored for unlabeled data, where the negative optimization is to explicitly regulate our EDL detector to output K low evidence values for a given outlier sample. However, negative optimization can not directly apply to inliers in unlabeled data.
Therefore, we derive an adaptive loss weight for encouraging the model to separately treat inliers and outliers in the unlabeled data in the learning process. It can largely avoid the interference between the ID and OOD features as well as the leading problem of uncertainty availability reduction.
Specifically, we achieve this by using the Fisher information matrix (FIM) introduced by $\mathcal{I}$-EDL \cite{deng2023uncertainty} to identify the amount of information for each class of unlabeled samples, while imposing smaller constraints on the confident classes with high evidence. Therefore, such optimization can sufficiently compress evidence values of unlabeled outliers without interfering with the learning of unlabeled inliers. 

To our knowledge, we are the first to exploit evidential deep learning in Open-set SSL and propose a novel Adaptive Negative Evidential Deep Learning (ANEDL) framework. Our main contributions can
be summarized as follows:
\begin{itemize}
\item We introduce EDL as an outlier detector for Open-set SSL, which enables quantification of different types of uncertainty and design of different uncertainty metrics for self-training and inference, respectively.
\item We propose a novel adaptive negative optimization strategy, which tailors EDL on the unlabeled dataset containing both inliers and outliers.
\item Extensive experiments on four public datasets, including CIFAR-10, CIFAR-100, ImageNet-30, and Mini-ImageNet, show that our method outperforms state-of-the-art methods in different settings.
\end{itemize}

\section{Related Works}

\subsection{Semi-supervised Learning}
Semi-supervised learning (SSL) aims to ease deep networks' reliance on massive and expensive labeled data by utilizing limited labeled data together with a large amount of unlabeled data. Existing works can be generally categorized into two paradigms: consistency regularization and pseudo labeling \cite{lee2013pseudo}. Consistency regularization assumes that the network should be invariant to different perturbations of the same instance. Pseudo-labeling takes confidently pseudo-labeled unlabeled data as labeled data to train the network. And many pseudo labeling methods like FixMatch \cite{sohn2020fixmatch}, FlexMatch \cite{zhang2021flexmatch} and AdaMatch \cite{berthelot2021adamatch} focus on how to select confidently pseudo-labeled data. Although pseudo-labeling methods have achieved tremendous success in SSL, they are deeply rooted in the close-set assumption and can not be directly applied to open-set SSL \cite{yu2020multi, saito2021openmatch}. Because noisy pseudo-labeled outliers may degrade the performance of self-supervised training.
On the other hand, OpenMatch \cite{saito2021openmatch} proposes that soft consistency regularization by minimizing the distance between its predictions on two augmentations of the same image can help SSL methods get rid of the close-set assumption can be kept for open-set SSL. Following OpenMatch, our method also applies soft consistency regularization to the outlier detector.

\subsection{Open-set Semi-supervised Learning}
Open-set SSL \cite{guo2020safe, chen2020semi} aims to detect outliers while categorizing inliers into correct close-set classes during training and inference. Existing open-set SSL methods recognize the weakness of SSL methods in terms of detecting outliers and they develop an extra outlier detector to exclude outliers from self-supervised training. Their outlier detectors are based on binary classifier activated by softmax function. For example, MTC \cite{yu2020multi} deploys a binary classifier to predict the probability of the instance belonging to outliers and updates the network parameters and the outlier score alternately. OpenMatch \cite{saito2021openmatch} takes one-vs-all (OVA) classifier as its outlier detector and proposes to apply soft consistency regularization to the outlier detector. An OVA classifier is composed of K binary classifiers and each classifier predicts the probability of the sample belonging to a specific class. 
However, softmax-based networks are notorious for inflating the probabilities of predicted classes \cite{szegedy2016rethinking, guo2017calibration, wilson2020bayesian}, and binary detectors perform poorly on tasks with a large number of classes \cite{carbonneau2018multiple}. To address these limitations, we adopt evidential deep learning (EDL) \cite{sensoy2018evidential} as the outlier detector for Open-set SSL.

\subsection{Evidential Deep Learning}
Evidential neural networks \cite{sensoy2018evidential} model the outputs as evidence to quantify belief masses and uncertainty based on the Dempster-Shafer Theory of Evidence (DST) \cite{sentz2002combination} and subjective logic \cite{josang2016subjective}. Similar to this work, \cite{malinin2018predictive} propose Prior Networks that explicitly consider the distributional uncertainty to quantify the mismatch of ID and OOD distribution. Compared with the standard neural network classifiers direct output the probability distribution of each sample, EDL obtains the density of classification probability assignments by parameterizing the Dirichlet distribution. Therefore, EDL can use the properties of Dirichlet distribution to distinguish different types of uncertainties, which shows an impressive performance in uncertainty quantification and is widely used in various applications. For example, \cite{zhao2020uncertainty} proposes a multi-source uncertainty framework combined with DST for semi-supervised node classification with GNNs. \cite{soleimany2021evidential} introduces evidential priors over the original Gaussian likelihood function to model the uncertainty of regression networks. \cite{bao2021evidential, bao2022opental} propose a general framework based on EDL for Open Set Recognition (OSR) and Open Set Temporal Action Localization (OSTAL), respectively. Compared with previous efforts, our work is the first to exploit EDL to Open-set SSL. For Open-set SSL, in addition to needing to distinguish unknown and known classes at test time, more importantly is how to better utilize the unlabeled data during training to learn the features of known classes.

\section{Method}

\subsection{Problem Setting}
Open-set semi-supervised learning (Open-set SSL) aims to learn a classifier by using a set of labeled and unlabeled datasets. Unlike traditional semi-supervised learning (SSL), the unlabeled dataset of Open-set SSL contains both ID and OOD samples, i.e., inliers and outliers. More specifically, for a K-way classification problem, let $\mathcal{D}_{l}=\{(\boldsymbol{x}^l_j, y^l_j)\}_{j=1}^{N_l}$ be the labeled dataset, containing $N_l$ labeled samples randomly generated from a latent distribution $\mathcal{P}$. The labeled example $\boldsymbol{x}^l_j$ can be classified into one of K classes, meaning that its corresponding label $y^l_j \in \{c_1, \cdots, c_K\}$. Let $\mathcal{D}_{u}= \{\boldsymbol{x}^i_j\}_{j=1}^{N_i} \cup \{\boldsymbol{x}^o_j\}_{j=1}^{N_o}$ be the unlabeled dataset consisting of $N_i + N_o$ examples with labels either from the K known classes (inliers) or never been seen (outliers). The goal of Open-set SSL is to detect outliers while classifying inliers into the correct class.

\subsection{Adaptive Negative Evidential Deep Learning}

\noindent\textbf{Approach Overview.} 
As shown in Figure \ref{fig:anedl}, our proposed framework contains three components: 
(1) a shared feature extractor $F(\cdot)$;
(2) a Softmax head $C(\cdot)$ that classifies samples by outputting K-way probabilities;
(3) an EDL head $D(\cdot)$ that quantifies uncertainty by outputting K evidence values to detect outliers in unlabeled data.
To adopt our framework to Open-set SSL, we propose an Adaptive Negative Optimization strategy to train EDL head by using all labeled and unlabeled samples. We introduce negative optimization to explicitly regularize our EDL detector to output low evidence values for uncertain classes, while deriving an adaptive loss weight to encourage the detector to pay more attention to these uncertain classes. For unlabeled data with confident prediction, we will use them jointly to train the Softmax classifier with labeled data.

\noindent\textbf{Evidential Outlier Detector.}
Traditional outlier detectors for Open-set SSL use Softmax activation function to predict confidence scores and achieve impressive performance. However, such softmax-based detectors only provide a point estimation of outlier probabilities, unable to quantify different types of uncertainty. To tackle the limitations of Softmax, we introduce evidential deep learning (EDL) \cite{sensoy2018evidential} to our framework as an outlier detection module. EDL proposes to integrate deep classification networks into the evidence framework of subjective logic \cite{josang2016subjective}, to jointly predict classification probabilities and quantify uncertainty.
Specifically, for K-way classification, EDL uses a network $g(\boldsymbol{\theta})$ to calculate its evidence vector $\boldsymbol{e}= g(\boldsymbol{x}|\boldsymbol{\theta})$. Note that an activation function, e.g. Relu or Softplus, is on top of $g(\boldsymbol{\theta})$ to guarantee non-negative evidence. Then EDL correlates its derived evidence vector $\boldsymbol{e}\in \mathbb{R}_{+}^{K}$ to the concentration parameter $\boldsymbol{\alpha} \in \mathbb{R}_{+}^{K}$ of Dirichlet distribution $D(\boldsymbol{p}|\boldsymbol{\alpha})$ with the equality ${\alpha}_i = e_i + 1$. 
Dirichlet distribution parameterized by $\boldsymbol{\alpha}$ models the density of classification probability assignments and uncertainty.
The expected probability of the $k^{th}$ class and uncertainty can be derived by
$$
{\hat{p}}_k = \frac{{\alpha}_k}{\alpha_{0}} \quad \text{and} \quad u = \frac{K}{\alpha_{0}}, 
$$
where 
$\alpha_{0}=\sum_{k=1}^K {\alpha}_k$. 
Higher values of $\alpha_{0}$ lead to 
more confident distributions.
Given a sample, ${\alpha}_k$ is incremented to update the Dirichlet distribution of the sample when evidence of $k^{th}$ class is observed.

\noindent\textbf{Joint Optimization with Softmax and EDL.} Although EDL-related works \cite{sensoy2018evidential, deng2023uncertainty} are capable of jointly classifying and quantifying uncertainty, we empirically find its deficiency in optimizing a network to learn representation for classification. 
That is, the classification performance of the evidential network is lower than that of a network activated by Softmax and optimized with only cross-entropy loss.
In this way, we propose to take the strength of both Softmax and EDL and jointly optimize their losses. In our proposed framework, Softmax head takes charge of learning representation and predicting classification probabilities, while EDL is responsible for quantifying uncertainty. During training, with EDL excluding outliers, we adopt an SSL method (i.e., FixMatch following OpenMatch \cite{saito2021openmatch}) to our Softmax head to enhance representation quality and classification accuracy. We also utilize counterfactual reasoning and adaptive margins proposed in DebiasPL \cite{wang2022debiased} to remove the bias of pseudo label in FixMatch.

\noindent\textbf{Adaptive Negative Optimization.}
The core challenge of Open-set SSL is that unlabeled data contains novel categories that are not seen in labeled training data.
To enhance the separation between inliers and outliers, detectors should output low confidence scores for outliers, but traditional binary detectors do not sufficiently leverage this information for model regularization, which may result in suboptimal performance \cite{malinin2018predictive}. 
Inspired by negative learning \cite{ishida2017learning}, we propose Adaptive Negative Optimization (ANO) to avoid providing misinformation to the EDL detector by focusing on uncertain classes. 
In particular, we introduce the negative optimization to explicitly regulate our detector to output low evidence values for uncertain classes. Meanwhile, adaptive loss weights are proposed to encourage EDL detectors to pay more attention to uncertain classes during the learning process.

More specifically, we adopt the Fisher information matrix (FIM) introduced by $\mathcal{I}$-EDL \cite{deng2023uncertainty} to identify the amount of information contained in each class of each sample, while imposing explicit constraints on the informative uncertainty classes. 
$\mathcal{I}$-EDL proves that the evidence of a class exhibits a negative correlation with its Fisher information, i.e., the class with more evidence corresponds to less Fisher information, and shows an impressive performance in uncertainty quantification \cite{deng2023uncertainty}. Therefore, we propose to use FIM as an indicator to distinguish inliers from outliers to adaptively regulate our model to pay more attention to uncertain classes.
For those uncertain classes with less evidence in unlabeled samples, we impose explicit constraints to output K low evidence values through the Kullback-Leibler (KL) divergence term. 
The objective function for unlabeled data is defined as
\begin{equation}
\begin{aligned}
\label{obj_unlabel}
\min_{\boldsymbol{\theta}} \ \mathbb{E}_{\boldsymbol{x} \sim \mathcal{D}_{u}} \left[ D_{KL}( \mathcal{U} \Vert \mathcal{N}(\mathbb{E}_{Dir(\boldsymbol{\alpha})}[\boldsymbol{p}], \sigma^2\mathcal{I}(\boldsymbol{\alpha})^{-1})\right]
\end{aligned}
\end{equation}
where $\mathcal{U}=\mathcal{N}(\boldsymbol{1}/K, \lambda^2 \mI)$, $\boldsymbol{\alpha} = g(\boldsymbol{x}|\boldsymbol{\theta}) + \boldsymbol{1}$, and $\mathcal{I}(\boldsymbol{\alpha})$ denotes the FIM of $Dir(\boldsymbol{\alpha})$ (the closed-form expression is provided in Appendix A.1). 
Eq.(\ref{obj_unlabel}) can be understood through the lens of probabilistic graphical models, where the observed labels $\hat{\boldsymbol{y}}$ are generated from the Dirichlet distribution $\boldsymbol{p}$ with its parameter $\boldsymbol{\alpha}$ calculated by passing the input sample $\boldsymbol{x}$ through the network. Since our unlabeled data contains both novel and known classes, we assume $\hat{\boldsymbol{y}} \sim \mathcal{N}(\boldsymbol{p}, \sigma^2 \mathcal{I}(\boldsymbol{\alpha})^{-1} )$ to make the model pay more attention to uncertain classes. For the target variable $\boldsymbol{y}$, we expect its Dirichlet distribution $\boldsymbol{p}$ to obey a uniform distribution, i.e. $\boldsymbol{y} \sim \mathcal{N}(\boldsymbol{1}/K, \lambda^2 \mI)$. Finally, we use KL divergence to constrain the observed labels $\hat{\boldsymbol{y}}$ and the target variable $\boldsymbol{y}$ to obey the same distribution. Due to the analyzable nature of the Dirichlet distribution, Eq.(\ref{obj_unlabel}) can be simplified as
\begin{equation}
\begin{aligned}
\label{obj_nedl}
& \mathcal{L}_{j}^{\text{N-EDL}} = \sum_{k=1}^{K}  (\frac{1}{K} - \frac{\alpha_{j k}}{\alpha_{j 0}} )^2 \psi_{1}(\alpha_{j k}) - \lambda_1 \log |\mathcal{I}(\boldsymbol{\alpha}_j)|,
\end{aligned}
\end{equation}
where 
${\alpha}_{j0} = \sum_{k=1}^{K}{\alpha}_{j k}$, and $\psi_{1}(\cdot)$ represents the \textit{trigamma} function, defined as $\psi_{1}(x) = d \psi(x) / dx = d^2 \ln \Gamma(x) / dx^2 $ (See Appendix A.2 for the detailed derivation process). We name the above weighted loss function as adaptive negative EDL loss. Since $\psi_{1}(x)$ is a monotonically decreasing function when $x > 0$, class labels with less evidence would be subject to greater penalties to achieve a flatter output. Conversely, once a certain class of evidence is learned, the weight will be reduced (see Fig.\ref{fig:anedl}), so that the inlier features in the unlabeled dataset do not interfere too much with the learning of outlier features. We also use labeled dataset to enhance the learning of inlier features. We follow $\mathcal{I}$-EDL \cite{deng2023uncertainty} to impose constraints on the uncertain class of labeled samples, outputting a sharp distribution that fits the ground truth.
The loss function for an labeled sample $(\boldsymbol{x}^l_j, \boldsymbol{y}^l_j) \in \mathcal{D}_l$ can be expressed as:
\begin{equation}
\begin{aligned}
\label{obj_pedl}
\mathcal{L}_{j}^{\text{P-EDL}}=&\sum_{k=1}^{K} \left[ (y_{j k}-\frac{\alpha_{j k}}{\alpha_{j 0}} )^2+\frac{\alpha_{j k}(\alpha_{j 0}- \alpha_{j k})}{\alpha_{j 0}^2(\alpha_{j 0} + 1)} \right] \psi_{1}(\alpha_{j k}) \\
&-\lambda_2 \log |\mathcal{I}(\boldsymbol{\alpha}_j)|,
\end{aligned}
\end{equation}
where $\boldsymbol{y}^l_j$ denotes one-hot encoded ground-truth of $\boldsymbol{x}^l_j$.


\noindent\textbf{Strengthened KL Loss.} 
KL divergence loss $\mathcal{L}_{j}^{\text{KL-ORI}} $ as part of classical EDL loss, aims to penalize evidence for misleading classes that a sample does not belong to: 
\begin{equation}
\label{obj_kl_ori}
\mathcal{L}_{j}^{\text{KL-ORI}} 
= D_{KL}(Dir(\boldsymbol{p}_j | \hat{\boldsymbol{\alpha}_j}) \Vert Dir(\boldsymbol{p}_j | \boldsymbol{1})).
\end{equation}
where $\hat{\boldsymbol{\alpha}}_j = \boldsymbol{\alpha}_j \odot(1 - \boldsymbol{y}^l_j) + \boldsymbol{y}^l_j$.
However, $\mathcal{L}_{j}^{\text{KL-ORI}} $ only shrinks misleading evidence but leaves non-misleading evidence ignored. To help enhance non-misleading evidence, we propose to utilize a strengthened KL-divergence to force our predicted Dirichlet distribution $Dir( \boldsymbol{p}_j | {\boldsymbol{\alpha}}_j)$ to approach a new target Dirichlet distribution $Dir(\boldsymbol{p}_j | \boldsymbol{\beta})$. In particular, we introduce a new KL divergence term as follows,
\begin{equation}
\label{obj_kl}
\mathcal{L}_{j}^{\text{KL}} 
= D_{KL}(Dir(\boldsymbol{p}_j | \boldsymbol{\alpha}_j) \Vert Dir(\boldsymbol{p}_j | \boldsymbol{\beta})). \\
\end{equation}
For labeled data, we set $\boldsymbol{\beta} = [1, \cdots, P, \cdots, 1] \in \mathbb{R}^{K}_+$, where $P$ is a large target evidence value for ground-truth, e.g. 100. 
For unlabeled data, we set $\boldsymbol{\beta} = \boldsymbol{1}$ as the target distribution. The detailed closed-form expression of $\mathcal{L}_{j}^{\text{KL}}$ can be found in Appendix A.3.

\noindent\textbf{Overall Loss Function.} 
For the EDL outlier detector, except Eq. (\ref{obj_nedl}), (\ref{obj_pedl}), (\ref{obj_kl}), we add a consistency loss $\mathcal{L}_\text{CON}$ to enhance the smoothness of detector, which is defined as
\begin{equation}
\begin{aligned}
\mathcal{L}_\text{CON}(\mathcal{D}_{u}) = \frac{1}{N_i + N_o} \sum_{j=1}^{N_i + N_o} \|\boldsymbol{\alpha}_{j}^{s} - \boldsymbol{\alpha}_{j}^{w}\|^2_2,
\end{aligned}
\end{equation}
where $\boldsymbol{\alpha}_{j}^{s}, \boldsymbol{\alpha}_{j}^{w}$ represent the predicted evidence vector of strong and weak augmentations of the same unlabeled sample. For the Softmax classifier, we compute the standard cross-entropy loss $\mathcal{L}_\text{CE}$ for labeled data and the FixMatch loss $\mathcal{L}_\text{FM}$ for the high-certainty inliers $\mathcal{I}_{u}$ identified by the EDL detector.
Therefore, the overall loss function of our proposed ANEDL can be expressed as
\begin{equation}
\begin{aligned}
\label{obj_final}
\mathcal{L}_\text{ANEDL} = &\mathcal{L}_\text{CE}(\mathcal{D}_{l}) + \mathcal{L}_\text{FM}(\mathcal{I}_{u}) \\
+ & \mathcal{L}_\text{ANO}(\mathcal{D}_{l}, \mathcal{D}_{u}) + \lambda_\text{CON} \mathcal{L}_\text{CON}(\mathcal{D}_{u}),
\end{aligned}
\end{equation}
where 
\begin{equation}
\begin{aligned}
\mathcal{L}_\text{ANO}(\mathcal{D}_{l}, \mathcal{D}_{u}) = 
& \lambda_\text{P-EDL} \frac{1}{N_l} \sum_{j=1}^{N_l} (\mathcal{L}_j^{\text{P-EDL}} + \mathcal{L}_{j}^{\text{KL}}) \\
+ & \lambda_\text{N-EDL} \frac{1}{N_i + N_o} \sum_{j=1}^{N_i + N_o} (\mathcal{L}_j^{\text{N-EDL}} + \mathcal{L}_{j}^{\text{KL}}),
\end{aligned}
\end{equation}
$\lambda_\text{P-EDL}, \lambda_\text{N-EDL}$ and $\lambda_\text{CON}$ are the hyperparameters used to control the trade-off for each objective.

To minimize our overall loss, we train our model for two stages. In the first stage, we pre-train our model with $\mathcal{L} = \mathcal{L}_\text{CE}(\mathcal{D}_{l}) +  \mathcal{L}_\text{ANO}(\mathcal{D}_{l}, \mathcal{D}_{u}) + \lambda_\text{CON} \mathcal{L}_\text{CON}(\mathcal{D}_{u})$ which drops $\mathcal{L}_\text{FM}$ for $E_{FM}$ epochs. In second stage, we start self-training of classifier and use $\mathcal{L}_\text{ANEDL}$ as loss function. Before every self-training epoch, we calibrate softmax head and EDL head to calculate uncertainty metric to select the inliers for softmax head to conduct self-training. 
\begin{table*}[t]
\begin{center}
\resizebox{\textwidth}{!}{
\begin{tabular}{clccclcclcclc}
\toprule
Dataset                  &  & \multicolumn{3}{c}{CIFAR-10}                  &  & \multicolumn{2}{c}{CIFAR-100}             &  & \multicolumn{2}{c}{CIFAR-100}             &  & ImageNet-30         \\ \cmidrule{1-1} \cmidrule{3-5} \cmidrule{7-8} \cmidrule{10-11} \cmidrule{13-13} 
Inlier/Outlier Classes &  & \multicolumn{3}{c}{6/4}                       &  & \multicolumn{2}{c}{55/45}                 &  & \multicolumn{2}{c}{80/20}                 &  & 20/10               \\ \cmidrule{1-1} \cmidrule{3-5} \cmidrule{7-8} \cmidrule{10-11} \cmidrule{13-13} 
No. of labeled samples   &  & 50         & 100        & 400                 &  & 50                  & 100                 &  & 50                  & 100                 &  & 10\%                \\ 
\midrule
FixMatch  
&  & 56.1$\pm$0.6 & 60.4$\pm$0.4 & 71.8$\pm$0.4          &  & 72.0$\pm$1.3          & 79.9$\pm$0.9          &  & 64.3$\pm$1.0          & 66.1$\pm$0.5          &  & 88.6$\pm$0.5          \\
MTC  
&  & 96.6$\pm$0.6 & 98.2$\pm$0.9 & 98.9$\pm$0.1         &  & 81.2$\pm$3.4          & 80.7$\pm$4.6          &  & 79.4$\pm$1.0          & 73.2$\pm$3.5          &  & 93.8$\pm$0.8          \\
T2T 
&  & 43.1$\pm$8.8 & 43.1$\pm$14.1 & 56.2$\pm$1.4         &  & 60.4$\pm$8.9          & 62.2$\pm$4.4          &  & 74.2$\pm$6.3          & 65.4$\pm$13.4          &  & 55.7$\pm$10.8        \\
OpenMatch 
&  & \textbf{99.3$\pm$0.9} & \textbf{99.7$\pm$0.2} & 99.3$\pm$0.2         &  & 87.0$\pm$1.1          & 86.5$\pm$2.1          &  & 86.2$\pm$0.6          & 86.8$\pm$1.4          &  & 96.4$\pm$0.7          \\ 
\midrule
Ours                     &  &  98.5$\pm$0.2          &   99.0$\pm$0.2         & \textbf{99.4$\pm$0.1} &  & \textbf{90.3$\pm$0.1} & \textbf{91.1$\pm$0.4} &  & \textbf{89.2$\pm$0.5} & \textbf{91.1$\pm$1.0} &  & \textbf{96.6$\pm$0.3} \\ 
\bottomrule
\end{tabular}}
\caption{Mean and standard deviation of AUROC (\%) on CIFAR-10, CIFAR-100 and ImageNet-30. Higher is better. For CIFAR-10 and CIFAR-100, the three runs correspond to three different folds. For ImageNet-30, the three runs are conducted on the same folder but with different random seeds.}
\label{tab:auroc}
\end{center}
\end{table*}

\begin{table*}[t]
\begin{center}
\resizebox{\textwidth}{!}{
\begin{tabular}{clccclcclcclc}
\toprule
Dataset                  &  & \multicolumn{3}{c}{CIFAR-10}          &  & \multicolumn{2}{c}{CIFAR-100} &  & \multicolumn{2}{c}{CIFAR-100} &  & ImageNet-30 \\ \cmidrule{1-1} \cmidrule{3-5} \cmidrule{7-8} \cmidrule{10-11} \cmidrule{13-13} 
Inlier/Outlier Classes &  & \multicolumn{3}{c}{6/4}               &  & \multicolumn{2}{c}{55/45}     &  & \multicolumn{2}{c}{80/20}     &  & 20/10       \\ \midrule
No. of labeled samples   &  & 50         & 100        & 400         &  & 50            & 100           &  & 50            & 100           &  & 10\%        \\ \midrule
FixMatch     
&  & 43.2$\pm$1.2 & 29.8$\pm$0.6 & 16.3$\pm$0.5  &  & 35.4$\pm$0.7    & 27.3$\pm$0.8    &  & 41.2$\pm$0.7    & 34.1$\pm$0.4    &  & 12.9$\pm$0.4  \\
MTC             
&  & 20.3$\pm$0.9 & 13.7$\pm$0.9 & 9.0$\pm$0.5 &  & 33.5$\pm$1.2    & 27.9$\pm$0.5    &  & 40.1$\pm$0.8    & 33.6$\pm$0.3    &  & 13.6$\pm$0.7  \\
T2T           
&  & 10.4$\pm$1.2 & 11.3$\pm$1.3 & 9.5$\pm$1.4 &  & 29.1$\pm$0.9    & 27.0$\pm$0.4    &  & 35.7$\pm$1.4    & 31.6$\pm$1.0    &  & 12.2$\pm$0.9 \\
OpenMatch     
&  & 10.4$\pm$0.9 & 7.1$\pm$0.5  & 5.9$\pm$0.5 &  & 27.7$\pm$0.4    & 24.1$\pm$0.6    &  & 33.4$\pm$0.2    & 29.5$\pm$0.3    &  & \textbf{10.4$\pm$1.0}  \\ \midrule
Ours                     &  &   \textbf{9.6$\pm$0.4}         &  \textbf{7.0$\pm$0.4}          & \textbf{5.5$\pm$0.2}   &  & \textbf{26.7$\pm$0.2}    & \textbf{23.2$\pm$0.6}    &  & \textbf{32.3$\pm$0.3}    & \textbf{28.5$\pm$0.2}    &  & 11.3$\pm$0.4  \\ \bottomrule
\end{tabular}}
\caption{Error rate (\%) corresponding to Table \ref{tab:auroc}. Lower is better.}
\label{tab:errorrate}
\end{center}
\end{table*}

\subsection{Uncertainty Metric for Open-set SSL}
\label{method:unc}
\noindent\textbf{Inlier Selection in Self-training.} To help better enhance recognition performance with unlabeled data, we aim to select inliers with accurate pseudo labels from unlabeled data for self-training. In FixMatch, pseudo label is generated by Softmax head and will also be used to softmax head as a supervision signal. Therefore, we propose to 
leverage EDL head to quantify the uncertainty of the class predicted by softmax head
 and select inliers with 

confident prediction which are more likely to be accurately pseudo-labeled. 

Specifically, given an unlabeled sample $\boldsymbol{x}_u$, the Softmax head  produce a one-hot pseudo label $\hat{\boldsymbol{y}} \in \{0,1\}^{K}$, the EDL head output evidence $\boldsymbol{\alpha} \in \mathbf{R}^{K}_{+}$. Then we use 
$$\text{M}_\text{Self-training} = \hat{\boldsymbol{y}} \cdot \boldsymbol{\alpha}$$
as the confidence value for $\boldsymbol{x}_u$. Larger values mean higher confidence, and we select top-$O$ most certain unlabeled samples as inliers. Here, $O$ is a hyperparameter.

\noindent\textbf{Outlier Detection for Inference.}
Detecting outliers in test data is also one of the tasks of Open-set SSL. 
Unlike selecting inliers from unlabeled data, detected inliers in test data do not have to consider calibrating with Softmax. Therefore, 
the uncertainty metric for inference is to accurately quantify epistemic uncertainty for better separation between inliers and outliers.
Based on the evidence vector $\boldsymbol{\alpha}$ predicted by EDL head, we can derive various uncertainty metrics, i.e. mutual information, differential entropy, and total evidence.

Previous EDL methods 
demonstrate total evidence 
${\alpha}_{0} = \sum_{k=1}^{K}{\alpha}_{k}$
achieves the best outlier detection performance \cite{sensoy2018evidential, deng2023uncertainty}. 
However, we empirically find that classes with small evidence values may form a long-tail effect because the EDL detector assigns a non-negative evidence value to each class. Although these evidence values are small, their total amount is not negligible, especially when dealing with tasks with a large number of classes.
To avoid these long-tailed evidence values disturbing the separation between inliers and outliers, we propose to sort the evidence vectors in descending order and use only the sum of the top-M evidence values as the confidence value for inference, which is defined as
\begin{equation}
\label{eq:uncertainty}
    \text{M}_\text{Inference} = \sum_{i=1}^{M} \bar{\alpha}_i,
\end{equation}
where $\bar{\boldsymbol{\alpha}}$ is the descending order of $\boldsymbol{\alpha}$.

\section{Experiments}

\subsection{Datasets}
We evaluate our method on four datasets, including CIFAR-10, CIFAR-100 \cite{krizhevsky2009learning}, ImageNet-30 \cite{hendrycks2016baseline} and Mini-ImageNet \cite{vinyals2016matching}. Appendix C.1 provides details about the dataset. We follow OpenMatch \shortcite{saito2021openmatch} to tailor the dataset for Open-set SSL. Given a dataset, we first split its class set into inlier and outlier class sets. Next, we randomly select inliers to form labeled, unlabeled, validation, and test dataset. Then all outliers will be randomly sampled into unlabeled and test datasets. For the four datasets, we set several experimental settings with different numbers of labeled samples and different amounts of inliers/outliers classes.

\subsection{Implementation Details}
 We employ a randomly initialized Wide ResNet-28-2 \cite{zagoruyko2016wide} to conduct experiments on CIFAR-10 and CIFAR-100, and a randomly initialized ResNet-18 \cite{he2016deep} for experiments on ImageNet-30 and Mini-ImageNet. The Softmax head is a linear layer and the EDL head is non-linear, which is composed of 4 MLP layers. For the pre-training stage, we set its length $E_{FM}$ as 10 for all experiments.
 We set the length of an epoch as 1024 steps, which means we select a new set of inliers every 1024 steps during self-training. For hyper-parameters of FixMatch, we set them the same as in OpenMatch \cite{saito2021openmatch}. We adopt SGD optimizer with 0.0001 weight decay and 0.9 momentum to train our model by setting initial learning rate as 0.03 with the cosine decrease policy. We set the batch size of labeled and unlabeled samples as 64 and 128 respectively for all experiments. Experiments on ImageNet-30 are conducted with a single 24-GB GPU and other experiments are conducted with a single 12-GB GPU. All experiments in this work are conducted over three runs with different random seeds, and we report the mean and standard derivation of three runs.

\begin{figure*}[t]
	\centering
	\includegraphics[width=0.99\textwidth]{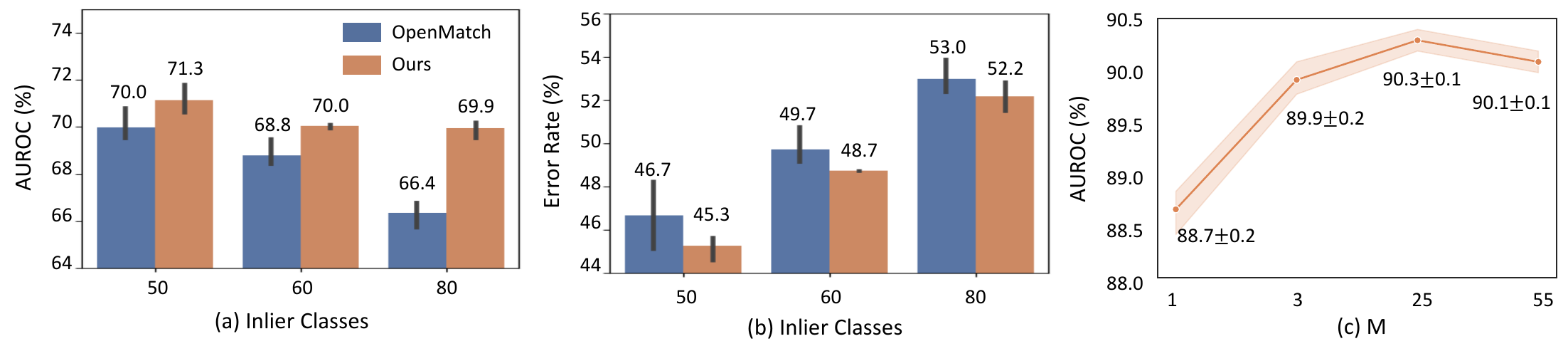}
	\caption{(a) AUROC ($\%$) and (b) Error rate ($\%$) compared to OpenMatch on Mini-ImageNet. (c) The effect of $M$ using in $\text{M}_\text{Inference}$ during inference. We use the same fold to conduct three runs and report the mean of three runs at the top of each bar. }
	\label{fig:exp_fig}
\end{figure*}

\subsection{Comparison with Other Methods}

To illustrate the effectiveness of our method on Open-set SSL,
we compare it against several baseline  
methods, including FixMatch \cite{sohn2020fixmatch}, MTC \cite{yu2020multi}, T2T \cite{huang2021trash} and OpenMatch \cite{saito2021openmatch}. 
We use classification error rate to evaluate the performance of classifier in inliers
and AUROC to evaluate the performance of outlier detection. AUROC is the general evaluation metric for novelty detection \cite{hendrycks2016baseline}. 
Table \ref{tab:auroc} and Table \ref{tab:errorrate} show the AUROC and error rate on the CIFAR-10, CIFAR-100, and ImageNet-30, respectively, where the results of other methods are quoted from OpenMatch \cite{saito2021openmatch}. Figure \ref{fig:exp_fig} (a) and (b) reports the experimental results on Mini-ImageNet, where the results of OpenMatch are obtained based on the author's own implementation.

As shown in Table \ref{tab:auroc}, we can see that our method outperforms the SOTA method, i.e., OpenMatch, in most experiment settings with respect to AUROC.
In Table \ref{tab:errorrate}, our proposed method also achieves SOTA performance in terms of error rate in 7 out of 8 settings. These results illustrate that our method is more advanced in epistemic uncertainty quantification and 
has superior performance in outlier detection as well as inlier classification.
Notably, in the more challenging settings with a large number of classes, i.e., settings on CIFAR-100, our method 
significantly outperforms OpenMatch in both error rate and AUROC, with improvements of $\sim \textbf{1\%}$ and $\sim \textbf{3\%}$. It shows our multinomial detector is more robust 
than the binary detector in identifying outliers from complex modalities when dealing with K-way classification tasks with large K. To further demonstrate our superiority in settings with a large number of inlier classes, we compare our method with OpenMatch 
on Mini-ImageNet, which is even more difficult than CIFAR-100 with larger image size and more complex patterns. 
As shown in Figure \ref{fig:exp_fig} (a) and (b), our method consistently outperforms OpenMatch with respect to both AUROC and error rate.


\subsection{Ablation Studies}
All experiments of our ablation studies are conducted under the setting of 55 inlier classes and 50 labeled samples per class on CIFAR-100.
We report the mean and standard deviation of three runs on three different folds.

\noindent\textbf{Component Analysis of ANO.}
Our ANO loss consists of two components, i.e. negative optimization and adaptive loss weight. For ``without NO and Adaptive'', we directly replace $\mathcal{L}_\text{ANO}$ to classical EDL loss, which means we apply no explicit constraints to outliers in unlabeled data. Since the information contained in unlabeled data keeps underutilized, 
both classification and outlier detection performance are suboptimal as shown in Table \ref{table:ano}.
For ``without Adaptive'', we remove our adaptive loss weight from $\mathcal{L}_\text{ANO}$, which means 
not using the FIM to identify the amount of information for each class of unlabeled samples. As a result, all unlabeled samples including inliers are equally regulated to output K-low evidence values. Although negative optimization compresses evidence values of unlabeled outliers, 
it simultaneously depresses the evidence of unlabeled inliers, making inlier features and outlier features interfere with each other. Therefore, its improvement compared to ``without NO and Adaptive'' is still limited. 
Our proposed $\mathcal{L}_\text{ANO}$, i.e., ``with NO and  Adaptive'', peaks the best performance by sufficiently compressing evidence values of unlabeled outliers without interfering with the learning of unlabeled inliers.
\begin{table}[t]
\begin{center}
\begin{tabular}{cc|cc}
\toprule
NO & Adaptive & AUROC & Error rate     \\ \midrule
     &          & 88.4$\pm$0.4 & 27.0$\pm$0.3\\
$\checkmark$     &          & 89.0$\pm$0.3 & 26.9$\pm$0.3\\
$\checkmark$     & $\checkmark$ & \textbf{90.3$\pm$0.1} & \textbf{26.7$\pm$0.2}\\ \bottomrule
\end{tabular}
\caption{Component analysis of Adaptive Negative Optimization (ANO).}
\label{table:ano}
\end{center}
\end{table}

\noindent\textbf{Evaluation of Uncertainty Metric.} 
To illustrate the effectiveness of designing different uncertainty metrics for self-training and inference, we conduct an ablation study and summarize it in Table \ref{table:metric}. Results show that using $\text{M}_\text{Self-training}$ and $\text{M}_\text{Inference}$ for self-training and inference, respectively, achieves the best performance in both AUROC and error rate. Since $\text{M}_\text{Self-training}$ cannot distinguish outliers and inliers with high aleatoric uncertainty, the AUROC using $\text{M}_\text{Self-training}$ is significantly lower than that of $\text{M}_\text{Inference}$. However, the effect of using $\text{M}_\text{Inference}$ during self-training outperforms using $\text{M}_\text{Self-training}$, probably due to the fact that inliers with high aleatoric uncertainty interfere with self-training. Note that using different metrics during inference only affects the effectiveness of outlier detection, not inlier classification. 
Furthermore, since $\text{M}_\text{Inference}$ uses the sum of top-$M$ evidence values as the metric, we also conduct an ablation study to evaluate the effect of $M$. As shown in Figure \ref{fig:exp_fig} (c), the sum of top $M=25$ evidence values is a better metric to detect outliers, compared with the sum of all $K=55$ evidence values. Note that the improvements steadily appear in each trained model. Comparison with other values of $M$ also shows that too small $M$ will miss useful evidence value and lead to insufficient measurement of uncertainty.

\begin{table}[t]
\begin{center}
\begin{tabular}{cc|cc}
\toprule
Self-training &  Inference    &    AUROC       &   Error rate             \\ \midrule
$\text{M}_\text{Inference}$ & $\text{M}_\text{Inference}$ &  89.8$\pm$0.2 &  27.8$\pm$0.2        \\ 
$\text{M}_\text{Self-training}$ & $\text{M}_\text{Self-training}$ &  88.4$\pm$0.3 &  26.7$\pm$0.2        \\ 
$\text{M}_\text{Self-training}$ & $\text{M}_\text{Inference}$   & \textbf{90.3$\pm$0.1} &  \textbf{26.7$\pm$0.2} \\ \bottomrule
\end{tabular}
\caption{Ablation study of uncertainty metric in self-training and inference. 
}
\label{table:metric}
\end{center}
\end{table}

\section{Conclusion}

In this work, we propose a novel framework, adaptive negative evidential deep learning (ANEDL), for open-set semi-supervised learning. ANEDL adopts evidential deep learning (EDL) to Open-set SSL for the first time and designs novel adaptive negative optimization method. In particular, EDL, as an advanced uncertainty quantification method, is deployed to estimate different types of uncertainty for inliers selection in self-training and outlier detection in inference. Furthermore, to enhance the separation between inliers and outliers, we propose adaptive negative optimization to explicitly compress the evidence value of outliers in unlabeled data and avoid interfering with the learning of inliers in unlabeled data with adaptive loss weight. Our extensive experiments on four datasets demonstrate that our method is superior to other SOTAs.

\section{Acknowledgments}
This work was supported by the National Key R$\&$D Program of China (2022YFE0200700), the Research Grants Council of the Hong Kong SAR, China (Project No. T45-401/22-N and Project No.14201620), the National Natural Science Foundation of China (Project No. 6237073934, 3234101132 62006219) and the Natural Science Foundation of Guangdong Province (2022A1515011579).

\bibliography{aaai24}
\appendix

\clearpage

\input{supplementary}

\end{document}

%% file: math_commands.tex
\usepackage{amsmath,amsfonts,bm}

\def\eqref#1{equation~\ref{#1}}

\def\1{\bm{1}}

\def\vtheta{{\bm{\theta}}}

\def\vb{{\bm{b}}}

\def\vp{{\bm{p}}}

\def\mI{{\bm{I}}}

\DeclareMathAlphabet{\mathsfit}{\encodingdefault}{\sfdefault}{m}{sl}
\SetMathAlphabet{\mathsfit}{bold}{\encodingdefault}{\sfdefault}{bx}{n}

\newcommand{\E}{\mathbb{E}}

\newcommand{\KL}{D_{\mathrm{KL}}}



%% file: supplementary.tex
\section*{A.Derivation}

This section provides some derivation presented in the main text, including the log determinant of the Fisher Information Matrix (FIM) of Dirichlet distribution, and the close-form expression of $\mathcal{L}_{j}^{\text{N-EDL}}$ and $\mathcal{L}_{j}^{\text{KL}}$ in Eq.(\ref{obj_nedl}) and Eq.(\ref{obj_kl}), respectively. 

\subsection*{A.1.FIM Derivation for Dirichlet Distribution}
\label{app_obj_fisher}

According to the definition of the Fisher Information Matrix (FIM), we can obtain the FIM of Dirichlet distribution as:
\begin{equation}
\begin{aligned}
\mathcal{I}(\boldsymbol{\alpha}) = \E \left[ \frac{\partial \ell}{\partial \boldsymbol{\alpha}} \frac{\partial \ell}{\partial \boldsymbol{\alpha}^{T}}\right] \in \mathbb{R}^{K \times K},
\end{aligned}
\tag{9}
\end{equation}
where $\ell = \log Dir(\vp|\boldsymbol{\alpha})$ is the log-likelihood function, and $Dir(\vp|\boldsymbol{\alpha})=\frac{\Gamma\left(\alpha_{0}\right)}{\prod_{k=1}^{K} \Gamma\left(\alpha_{k}\right)} \prod_{k=1}^{K} p_{k}^{\alpha_{k}-1}, \alpha_{0}=\sum_{k=1}^{K} \alpha_{k}$.
By applying Lemma 5.3 in \cite{lehmpointann2006theory}, the FIM can be expressed as $\mathcal{I}(\boldsymbol{\alpha}) = \E_{Dir(\vp|\boldsymbol{\alpha})} \left[ - \partial^2 \ell /  \partial \boldsymbol{\alpha} \boldsymbol{\alpha}^{T} \right]$. Hence, we can calculate each element by
$$
\begin{aligned}
 \relax[\mathcal{I}(\boldsymbol{\alpha})]_{i j} &= \E_{Dir(\vp|\boldsymbol{\alpha})} \left[ - \frac{\partial^2}{\partial \alpha_i \partial \alpha_j } \log Dir(\vp|\boldsymbol{\alpha}) \lvert \boldsymbol{\alpha} \right] \\
& = \E_{Dir(\vp|\boldsymbol{\alpha})} \left[ - \frac{\partial}{\partial \alpha_j } \left( \psi\left(\alpha_{0}\right) - \psi\left(\alpha_{i}\right) + \log p_{i} \right) \right] \\
& = \begin{cases}
\psi^{(1)}\left(\alpha_{i}\right) - \psi^{(1)}\left(\alpha_{0}\right), & i=j, \\ 
-\psi^{(1)}\left(\alpha_{0}\right), & i \neq j,
\end{cases}
\end{aligned}
$$
where $\Gamma(\cdot)$ is the \textit{gamma} function, $\psi(\cdot)$ is the \textit{digamma} function, $\psi^{(1)}(\cdot)$ is the \textit{trigamma} function, defined as $\psi^{(1)}(x) = d \psi(x) / dx = d^2 \ln \Gamma(x) / dx^2$. Then, we can get the matrix form of the FIM:
\begin{equation}
\begin{aligned}
\label{fisher_matrix}
\mathcal{I}(\boldsymbol{\alpha}) = &
\begin{bmatrix}
\psi^{(1)}\left(\alpha_{1}\right) - \psi^{(1)}\left(\alpha_{0}\right) & \cdots & - \psi^{(1)}\left(\alpha_{0}\right) \\
- \psi^{(1)}\left(\alpha_{0}\right) & \cdots & - \psi^{(1)}\left(\alpha_{0}\right) \\
\vdots & \ddots & \vdots \\
- \psi^{(1)}\left(\alpha_{0}\right) & \cdots & \psi^{(1)}\left(\alpha_{K}\right) - \psi^{(1)}\left(\alpha_{0}\right)
\end{bmatrix} \\
= & \ \text{diag}([\psi^{(1)}(\alpha_{1}), \cdots, \psi^{(1)}(\alpha_{K})])- \psi^{(1)}(\alpha_{0}) \boldsymbol{1}\boldsymbol{1}^T,
\end{aligned}
\tag{10}
\end{equation}
where $\boldsymbol{1} = [1; \cdots; 1] \in \mathbb{R}^{K}$.

Next, we derive the close-form of log determinant of Eq. (\ref{fisher_matrix}). Let $\vb=[ \psi^{(1)}(\alpha_{1}), \cdots, \psi^{(1)}(\alpha_{K})]^T$, by applying Matrix-Determinant Lemma, we have
\begin{equation}
\begin{aligned}
|\mathcal{I}(\boldsymbol{\alpha})| & = |\text{diag}(\vb)| \cdot (1 - \psi^{(1)}(\alpha_{0}) \boldsymbol{1}^T \text{diag}(\vb)^{-1}\boldsymbol{1}) \\
& = \prod_{k=1}^{K} \psi^{(1)}(\alpha_{k}) (1 - \sum_{j=1}^{K} \frac{\psi^{(1)}(\alpha_{0})}{\psi^{(1)}(\alpha_{j})}).
\end{aligned}
\tag{11}
\end{equation}
Therefore, we can obtain
\begin{equation}
\begin{aligned}
\label{log_det_fisher}
\log |\mathcal{I}(\boldsymbol{\alpha})| = \sum_{k=1}^{K} \log \psi^{(1)}(\alpha_{k}) + \log (1 - \sum_{k=1}^{K} \frac{\psi^{(1)}(\alpha_{0})}{\psi^{(1)}(\alpha_{k})}).
\end{aligned}
\tag{12}
\end{equation}

\subsection*{A.2.Simplification of $\mathcal{L}_{j}^{\text{N-EDL}}$ in Eq.(\ref{obj_nedl})}
\label{app_obj_nedl}

Since $\mathcal{L}_{j}^{\text{N-EDL}} = \KL(\mathcal{U} \Vert \mathcal{N}( \E_{Dir(\boldsymbol{\alpha}_{j})}[\vp], \sigma^2\mathcal{I}(\boldsymbol{\alpha}_{j})^{-1})$,
where $\mathcal{U} = \mathcal{N}(\boldsymbol{1}/K, \lambda^2 \mI)$.
Then, we have
$$
\begin{aligned}
& \KL( \mathcal{N}(\boldsymbol{1}/K, \lambda^2 \mI) \Vert \mathcal{N}( \E_{Dir(\boldsymbol{\alpha}_{j})}[\vp], \sigma^2\mathcal{I}(\boldsymbol{\alpha}_{j})^{-1}) \\
\propto & \frac{1}{2\sigma^2} (\frac{1}{K} - \frac{\boldsymbol{\alpha}_{j}}{\alpha_{j 0}})^T \mathcal{I}(\boldsymbol{\alpha}_{j})(\frac{1}{K} - \frac{\boldsymbol{\alpha}_{j}}{\alpha_{j 0}}) - \frac{1}{2} \log |\mathcal{I}(\boldsymbol{\alpha}_j)| \\
\propto & \sum_{k=1}^{K}(\frac{1}{K} - \frac{\alpha_{j k}}{\alpha_{j 0}})^2 \psi^{(1)}(\alpha_{j k}) - \lambda_1 \log |\mathcal{I}(\boldsymbol{\alpha}_j)|.
\end{aligned}
$$
The second formula is derived from the formula of KL divergence between two multivariate Gaussians, and the last formula is derived on the closed-form of $\mathcal{I}(\boldsymbol{\alpha}_{j})$ in Eq. (\ref{fisher_matrix}) 
$$
\begin{aligned}
& (\frac{1}{K} - \frac{\boldsymbol{\alpha}_{j}}{\alpha_{j 0}})^T \mathcal{I}(\boldsymbol{\alpha}_{j})(\frac{1}{K} - \frac{\boldsymbol{\alpha}_{j}}{\alpha_{j 0}}) \\
= & (\frac{1}{K} - \frac{\boldsymbol{\alpha}_{j}}{\alpha_{j 0}})^T \text{diag}([\psi^{(1)}(\alpha_{j1}), \cdots, \psi^{(1)}(\alpha_{jK})]) (\frac{1}{K} - \frac{\boldsymbol{\alpha}_{j}}{\alpha_{j 0}})  \\
& - (\frac{1}{K} - \frac{\boldsymbol{\alpha}_{j}}{\alpha_{j 0}})^T \psi^{(1)}(\alpha_{j0}) \boldsymbol{1}\boldsymbol{1}^T (\frac{1}{K} - \frac{\boldsymbol{\alpha}_{j}}{\alpha_{j 0}}) \\
= & \sum_{k=1}^{K}(\frac{1}{K} - \frac{\alpha_{j k}}{\alpha_{j 0}})^2 \psi^{(1)}(\alpha_{j k}),
\end{aligned}
$$
where the last formula comes from the fact that
$$
\begin{aligned}
& (\frac{1}{K} - \frac{\boldsymbol{\alpha}_{j}}{\alpha_{j 0}})^T \psi^{(1)}(\alpha_{j 0}) \boldsymbol{1}\boldsymbol{1}^T (\frac{1}{K} - \frac{\boldsymbol{\alpha}_{j}}{\alpha_{j 0}}) \\
= & \psi^{(1)}(\alpha_{j 0})\left( \sum_{i=1}^K (\frac{1}{K} - \frac{\alpha_{j i}}{\alpha_{j 0}})\right)^2 \\
= & \psi^{(1)}(\alpha_{j 0})\left( 1 - \sum_{i=1}^K \frac{\alpha_{j i}}{\alpha_{j 0}}\right)^2 = 0.
\end{aligned}
$$

\begin{figure}[t]
	\centering
	\includegraphics[width=0.46\textwidth]{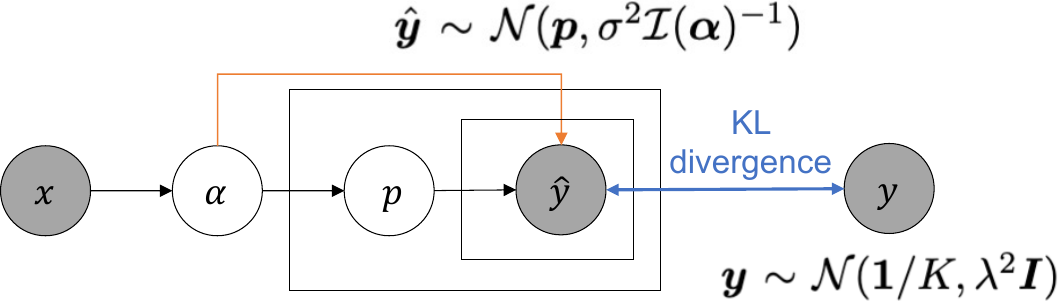}
	\caption{Graphical model representation of $\mathcal{L}^{\text{N-EDL}}$.}
	\label{fig:prob}
\end{figure}

Figure \ref{fig:prob} also provides the lens of probabilistic graphical models to understand $\mathcal{L}^{\text{N-EDL}}$.

\subsection*{A.3.Closed-form of $\mathcal{L}_{j}^{\text{KL}}$ in Eq. (5)}
\label{app_obj_kl}

Since $\mathcal{L}_{j}^{\text{KL}} 
= \KL(Dir(\vp_j | \boldsymbol{\alpha}_j) \Vert Dir(\vp_j | \boldsymbol{\beta})),$
where
$$
Dir(\vp_j|\boldsymbol{\alpha}_j)=\frac{1}{B(\boldsymbol{\alpha}_j)} \prod_{k=1}^{K} p_{j k}^{\alpha_{k}-1}, B(\boldsymbol{\alpha}_j) = \frac{\prod_{k=1}^{K} \Gamma\left(\alpha_{j k}\right)}{\Gamma\left(\sum_{k=1}^{K} \alpha_{j k}\right)}.
$$
Then, we have
\begin{equation}
\begin{aligned}
\label{app_kl}
\mathcal{L}_{j}^{\text{KL}} 
= & \KL(Dir(\vp_j | \boldsymbol{\alpha}_j) \Vert Dir(\vp_j | \boldsymbol{\beta})) \\
= & \E_{Dir(\vp_j | \boldsymbol{\alpha}_j)} \left[\log \frac{Dir(\vp_j | \boldsymbol{\alpha}_j)}{Dir(\vp_j | \boldsymbol{\beta}))} \right] \\
= & \E_{Dir(\vp_j | \boldsymbol{\alpha}_j)} \left[\log Dir(\vp_j | \boldsymbol{\alpha}_j) - \log Dir(\vp_j | \boldsymbol{\beta})) \right]
\\
= & \E_{Dir(\vp_j | \boldsymbol{\alpha}_j)} \left[ - \log B(\boldsymbol{\alpha}_j) + \sum_{k=1}^{K}(\alpha_{j k} - 1) \log p_{j k} \right] \\
& - \E_{Dir(\vp_j | \boldsymbol{\alpha}_j)} \left[ - \log B(\boldsymbol{\beta}) + \sum_{k=1}^{K}(\beta_{k} - 1) \log p_{j k}\right] \\
= & \log \frac{B(\boldsymbol{\beta})}{B(\boldsymbol{\alpha}_j)} + \E_{Dir(\vp_j | \boldsymbol{\alpha}_j)} \left[\sum_{k=1}^{K}(\alpha_{j k} - \beta_{k}) \log p_{j k} 
 \right].
\end{aligned}
\tag{13}
\end{equation}

Therefore, we obtain 
\begin{equation}
\begin{aligned}
\mathcal{L}_{j}^{\text{KL}} = & \log \Gamma(\alpha_{j 0}) - \sum_{k=1}^{K} \log \Gamma(\alpha_{j k}) \\
& - \log \Gamma(\beta_{0}) + \sum_{k=1}^{K} \log \Gamma(\beta_{k}) \\
&+ \sum_{k=1}^{K}(\alpha_{j k} - \beta_{k}) \left[\psi(\alpha_{j k}) - \psi(\alpha_{j 0}) \right],
\end{aligned}
\tag{14}
\end{equation}
where $\alpha_{j 0} = \sum_{k=1}^{K} \alpha_{j k}$, $\beta_{0} = \sum_{k=1}^{K} \beta_{k}$.

\begin{algorithm*}[t]
   \caption{Adaptive Negative Evidential Deep Learning}
   \label{alg:train}
    \begin{algorithmic}[]
   \STATE {\bfseries Input:} Labeled dataset $\mathcal{D}_{l}=\{(\boldsymbol{x}^l_j, y^l_j)\}_{j=1}^{N_l}$, unlabeled dataset $\mathcal{D}_{u}= \{\boldsymbol{x}^i_j\}_{j=1}^{N_i} \cup \{\boldsymbol{x}^o_j\}_{j=1}^{N_o}$, pseudo-inlier set $\mathcal{I}_{u} = \emptyset$, learning rate $\beta$, total epochs $E$, steps in an epoch $S$, trade-off parameters 
   ${\lambda}_\text{P-EDL}$, ${\lambda}_\text{N-EDL}$ and 
   ${\lambda}_\text{CON}$, initialized $\vtheta$
   \FOR{$e=1,\cdots, E$}
       \IF{$e > E_{FM}$}
        \STATE \textbf{Update} $\mathcal{I}_{u} = Select(\theta, \mathcal{D}_{u})$ \COMMENT{See Section: Uncertainty Metric for Open-set SSL}
        \ENDIF
        \FOR{$s=1,\cdots, S$}
            \STATE \textbf{Sample} a batch of labeled data $\mathcal{B}_{l} \in \mathcal{D}_{l}$ and unlabeled data $\mathcal{B}_{u} \in \mathcal{D}_{u}$
            \STATE \textbf{Compute} $\mathcal{L}_\text{ANO}(\mathcal{B}_{l}, \mathcal{B}_{u}) = \lambda_\text{P-EDL} (\mathcal{L}^{\text{P-EDL}}(\mathcal{B}_{l}) + \mathcal{L}^{\text{KL}}(\mathcal{B}_{l})) + \lambda_\text{N-EDL} (\mathcal{L}^{\text{N-EDL}}(\mathcal{B}_{u}) + \mathcal{L}^{\text{KL}}(\mathcal{B}_{u}))$ \COMMENT{See Eq.(8)}
            \STATE \textbf{Compute} $\mathcal{L} = \mathcal{L}_\text{CE}(\mathcal{B}_{l}) +  \mathcal{L}_\text{ANO}(\mathcal{B}_{l}, \mathcal{B}_{u}) + \lambda_\text{CON} \mathcal{L}_\text{CON}(\mathcal{B}_{u})$ \COMMENT{See Eq.(7)}
            \IF{$e > E_{FM}$}
            \STATE \textbf{Sample} a batch of pseudo-inliers $\mathcal{B}_{u}^{\mathcal{I}}\in \mathcal{I}_{l}$
            \STATE \textbf{Compute} $\mathcal{L} += \mathcal{L}_\text{FM}(\mathcal{B}_{u}^{\mathcal{I}})$
            \ENDIF
            \STATE $\vtheta \leftarrow \vtheta - \beta \nabla_\vtheta \mathcal{L}$ 
            \ENDFOR
    \ENDFOR
    \end{algorithmic}
\end{algorithm*}

\section*{B.Pseudo-code of ANEDL}
\label{app_pseudo_code}
The complete pseudo-code of ANEDL is outlined in Algorithm \ref{alg:train}.

\section*{C.Experimental Details}
\label{app_exp}

\subsection*{C.1.Datasets}
\noindent\textbf{CIFAR-10} \cite{krizhevsky2009learning} has 10 categories and each category has 6,000 images of size $32 \times 32$. 1,000 images of each category are used for test. We choose 6/4 classes as inlier/outlier classes. The amount of labeled images of each inlier class varies in $\{50,100,400\}$. And the number of validation images of each inlier class is 50.

\noindent\textbf{CIFAR-100} \cite{krizhevsky2009learning} is composed of 60,000 images of size $32 \times 32$  from 100 classes. 100 images of each class are used for test. We choose 55/80 classes as inlier classes, resulting in 45/20 outliers classes. The amount of labeled images of each inlier class varies in $\{50,100\}$. The number of validation images of each inlier class is 50.

\noindent\textbf{ImageNet-30} \cite{hendrycks2016baseline} is a subset of ImageNet \cite{deng2009imagenet}. It has 30 classes and each class contains 1,300 images of size $224 \times 224$. 100 images per class are used for test. Inlier class number is 20. The amounts of labeled and validation images per inlier class are both 130. 

\noindent\textbf{Mini-ImageNet} \cite{vinyals2016matching} is also a subset of ImageNet. It has 100 classes and each class contains 600 images of size $84 \times 84$. 100 images of each class are used for test. The number of inlier classes varies in $\{50,60,80\}$. The amounts of labeled and validation images of each inlier class are both 50.



\section*{D.Additional Experimental Results}

\subsection*{D.1.Evaluation on KL Divergence.}

Besides shrinking misleading evidence, we propose to boost non-misleading evidence with KL divergence loss to improve the separation of inliers and outliers. In Eq. (\ref{obj_kl}), we keep $\boldsymbol{\alpha}_i$ unchanged and setting target Dirichlet Distribution to $Dir(\boldsymbol{p}_i | \boldsymbol{\beta})$, where $\boldsymbol{\beta} = [1, \cdots, P, \cdots, 1] $. $\boldsymbol{{\beta}_{100}}$ and $\boldsymbol{{\beta}_{200}}$ denote $P=100$ and $P=200$, respectively. Table \ref{table:kl}
shows that our strengthen KL divergence can effectively improve the separation of inliers and outliers by boosting non-misleading evidence. However, a too large $P$ may cause an unstable training process, 
resulting in suboptimal performance.

\subsection*{D.2.Evaluation of DebiasPL}

As mentioned in ``Joint Optimization with Softmax and EDL '' of Section 3.2, we utilize DebiasPL \cite{wang2022debiased} to remove the bias of pseudo label in Fixmatch. In Table \ref{table:debias}, we study the influence of DebiasPL. Since OpenMatch is trained without DebiasPL, we remove DebiasPL from our method for a fair comparison with OpenMatch. We can see that, even trained without DebiasPL, our method still outperforms OpenMatch with respect to both AUROC and error rate. 

\subsection*{D.3.Case Study}
\begin{figure*}[t]
	\centering
	\includegraphics[width=0.98\textwidth]{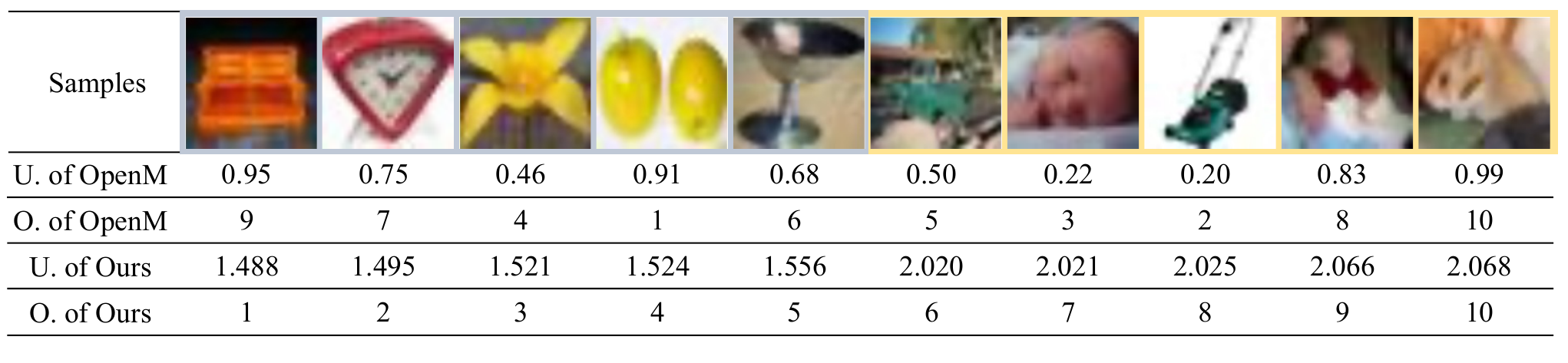}
	\caption{Case study. U. and O. represent uncertainty and order, respectively. The uncertainty of our method can better distinguish between inliers and outliers than that of OpenMatch. (Samples with blue and yellow backgrounds indicate inliers and outliers, respectively.)}
	\label{fig:case}
\end{figure*}

 We further conduct a case study under the setting of 55 inlier classes and 50 labeled samples per class on CIFAR-100, as shown in Figure \ref{fig:case}. 
 We derive uncertainty value of our method by Eq.(\ref{eq:uncertainty}), while the value of OpenMatch is derived from the outlier likelihood $p_{ood}$ predicted by its binary detector. We sort uncertainty values in descending order and report the order in Figure \ref{fig:case}. We can see that compared with OpenMatch, our method can better distinguish between inliers and outliers.

\begin{table}[t]
\begin{center}
\begin{tabular}{c|c|c}
\toprule
KL divergence & AUROC & Error rate     \\ \midrule
$D_{KL}(D(\boldsymbol{p}_i | \hat{\boldsymbol{\alpha}}_i) \Vert D(\boldsymbol{p}_i | \boldsymbol{1}))$& 89.9$\pm$0.3 & 27.0$\pm$0.2\\
$D_{KL}(D(\boldsymbol{p}_i | \boldsymbol{\alpha}_i) \Vert D(\boldsymbol{p}_i | \boldsymbol{{\beta}_{100}}))$& \textbf{90.3$\pm$0.1} &  \textbf{26.7$\pm$0.2} \\
$D_{KL}(D(\boldsymbol{p}_i | \boldsymbol{\alpha}_i) \Vert D(\boldsymbol{p}_i | \boldsymbol{{\beta}_{200}}))$ &  89.9$\pm$0.9        &     26.9$\pm$0.2       \\ \bottomrule
\end{tabular}
\caption{Evaluation on KL divergence.}
\label{table:kl}
\end{center}

\end{table}

\begin{table}[t]
\centering
\begin{tabular}{c|c|c}
\toprule
Method           & AUROC      & Error rate \\ \midrule
OpenMatch        & 87.0$\pm$1.1 & 27.7$\pm$0.4 \\ 
Ours(w/o DebiasPL) & 90.0$\pm$0.5 & 27.0$\pm$0.2 \\
Ours(w DebiasPL)   & \textbf{90.3$\pm$0.1} & \textbf{26.7$\pm$0.2} \\ \bottomrule
\end{tabular}
\caption{Evaluation on DebiasPL}
\label{table:debias}
\end{table}

